\pdfoutput=1
\documentclass[11pt]{article}

\usepackage[preprint]{acl}

\usepackage{times}
\usepackage{latexsym}

\usepackage[T1]{fontenc}
\usepackage[utf8]{inputenc}

\usepackage{microtype}

\usepackage{inconsolata}

\usepackage{amsmath}
\usepackage{amssymb}

\usepackage{graphicx}

\usepackage{arydshln}

\title{MoCE: Adaptive \underline{M}ixture \underline{o}f \underline{C}ontextualization \underline{E}xperts for Byte-based Neural Machine Translation}

\author{Langlin Huang \textsuperscript{\rm 1,3,4}\footnotemark[2], Mengyu Bu \textsuperscript{\rm 1,3}, Yang Feng \textsuperscript{\rm 1,2,3}\footnotemark[1] \\
  \textsuperscript{\rm 1} Key Laboratory of Intelligent Information Processing \\
  Institute of Computing Technology, Chinese Academy of Sciences \\
  \textsuperscript{\rm 2} Key Laboratory of AI Safety, Chinese Academy of Sciences \\
  \textsuperscript{\rm 3} University of Chinese Academy of Sciences \\
  \textsuperscript{\rm 4} Washington University in St. Louis \\
  \texttt{\href{mailto:h.langlin@wustl.edu}{h.langlin@wustl.edu}, \{\href{mailto:bumengyu23z@ict.ac.cn}{bumengyu23z}, \href{mailto:fengyang@ict.ac.cn}{fengyang}\}@ict.ac.cn} \\
  }

\begin{document}
\maketitle
\renewcommand{\thefootnote}{\fnsymbol{footnote}}
\footnotetext[1]{Corresponding author.}
\footnotetext[2]{This paper was done when Langlin Huang studied at Institute of Computing Technology, Chinese Academy of Sciences.}
\renewcommand{\thefootnote}{\arabic{footnote}}
\begin{abstract}
Byte-based machine translation systems have shown significant potential in massively multilingual settings. Unicode encoding, which maps each character to specific byte(s), eliminates the emergence of unknown words, even in new languages. This avoids out-of-vocabulary risk in multilingual translation and enables broad language scalability. However, byte-level tokenization results in sequences that are hard to interpret due to limited semantic information per byte. Local contextualization has proven effective in assigning initial semantics to tokens, improving sentence comprehension. Nevertheless, variations in encoding rules across languages necessitate an adaptive approach for effective contextualization. To this end, we propose Mixture of Contextualization Experts (MoCE), adaptively selecting and mixing attention heads, which are treated as contextualization experts. This enhances the flexibility of contextualization scales and allows models to search for better contextualization combinations. Experiment results show that our method outperforms existing methods without extensive manual adjustment of hyper-parameters and surpasses subword-based models with fewer parameters in Ted-59 dataset. Our code is available at \url{https://github.com/ictnlp/MoCE}.
\end{abstract}

\section{Introduction}
Neural Machine Translation (NMT) is a consistently hot research topic, and recent years have seen the growing significance of multilingual language modeling~\cite{DBLP:journals/corr/abs-2306-10968}. The selection of tokenization and vocabulary is critical to multilingual language models, which plays an important role in vectorization of texts and discretization of predicted hidden states. 
% While some models, such as NLLB~\cite{NLLB-DBLP:journals/corr/abs-2207-04672} and Llama3.1~\cite{dubey2024llama}, use large vocabularies to ensure word coverage, others like Llama2~\cite{LLAMA-2-DBLP:journals/corr/abs-2307-09288} and Mistral~\cite{Mistral-7b-DBLP:journals/corr/abs-2310-06825} opt for byte fallback strategy. 
While some models~\cite{NLLB-DBLP:journals/corr/abs-2207-04672,DBLP:journals/corr/abs-2407-21783} use large vocabularies to ensure word coverage, others~\cite{LLAMA-2-DBLP:journals/corr/abs-2307-09288,Mistral-7b-DBLP:journals/corr/abs-2310-06825} opt for byte fallback strategy.
These approaches allow them to completely avoid unknown words with a smaller vocabulary size. Byte-based models like \citet{DBLP:journals/tacl/XueBCANKRR22,DBLP:conf/nips/YuSFAZL23,shaham-levy-2021-neural} convert all words into UTF-8 byte, which further reduces the vocabulary size to about 256. This strategy also reduces the size of the embedding table, saving parameters and accelerating token embedding and inference. Besides, it eliminates the unknown-word problem and can be easily generalized to massively multilingual scenarios. Empirical study~\cite{edman-etal-2024-character} has also shown the performance superiority of byte-based MNT models.

However, the drawbacks of byte-based models are obvious, most notably that an individual byte struggles to convey a specific semantic meaning. Therefore, various contextualization methods~\cite{DBLP:journals/tacl/LeeCH17, DBLP:journals/tacl/ClarkGTW22} have been proposed to alleviate this problem. MEGABYTE~\cite{DBLP:conf/nips/YuSFAZL23} reassembles byte streams into groups of four, constructing group representations by concatenating their hidden states. CharFormer~\cite{tay2022charformer} and LOBEF~\cite{sreedhar-etal-2023-local} employ local-contextualization techniques to encode bytes, with CharFormer using mean-pooling and LOBEF using Convolutional Neural
Networks (CNNs). MSC~\cite{huang-feng-2024-integrating} argues that a byte should contribute to multiple neighboring contexts, necessitating a multi-scale contextualization approach. To this end, MSC groups hidden state dimensions and assigns CNNs with different kernel sizes to each group.

Although MSC provides an effective framework for modeling multi-scale contextualization and achieves state-of-the-art performance, it suffers from the limitation of manually predefined scales. This reduces the model's ability to generalize to multilingual scenarios, particularly in massively multilingual machine translation, which may involve over 50 languages. Under UTF-8 rule, a character may convert to 1 to 4 bytes, depending on the language. This leads to varying requirements of contextualization scale for different languages. 
However, once MSC decides the contextualization scales, they are unchangeable for any input.

To address this, we leverage the concept of Mixture of Experts (MoE)~\cite{shazeer2017} and propose Mixture of Contextualization Experts (MoCE), which can adaptively determine CNN kernel sizes based on each input text.
Specifically, we modify Multi-Head Attention to propose Adaptive MultiScale-Headed Attention (Ada-MSHA) module. This proposed attention allows each head to be locally contextualized and the contextualization scales are adaptive to the input. Instead of predefined scales that MSC uses, MoCE dynamically combines different scales with model needs. The flexibility of contextualization scales is therefore significantly enhanced, resulting in better performance. Additionally, we prove that given language ID as prior knowledge benefits the scale selection.

Experiment results on two massively multilingual translation datasets, Ted-59 and OPUS-100, demonstrate our proposed method outperforms other byte-based translation models with similar parameter use. Compared with the subword-based model, MoCE requires fewer parameters while performing better in Ted-59 dataset.

\section{Background}
\subsection{Mixture-of-Experts (MoE)}

MoE was designed mainly to increase the potential parameters of a model~\cite{shazeer2017,lepikhin2021gshard,DBLP:journals/corr/abs-2401-04088}. Recently, it is also used in multi-domain~\cite{DBLP:journals/corr/abs-2311-00285} or multi-task~\cite{DBLP:journals/corr/abs-2405-11530,huang2023lorahub} scenarios. 

An "expert" usually represents a layer within the model, and MoE provides multiple counterparts for the same layer. During computation, only one or a small number of experts are activated at a time, increasing model's modeling ability with limited extra computational cost.  It is worth noting that the counterparts are not necessarily the same structure; \citet{ramachandran2018diversity} uses heterogeneous experts, such as CNNs with different kernel sizes.

In terms of expert selection, a typical method is predicting the selection probability distribution of the experts and then choosing the $k$ most possible experts~\cite{shazeer2017,ramachandran2018diversity}. As shown in \eqref{eq:moe1}, $x$ and $y$ are the input and output of an MoE layer respectively. $E_i$ represents the $i^{th}$ expert, and $G_i$ is the normalized probability of choosing the $i^{th}$ expert, given by \eqref{eq:moe2}.
% Such process is named routing, where the module is a router.

\begin{equation}
\label{eq:moe1}
    y=\sum_i^nG_i(x)E_i(x)
\end{equation}

\begin{equation}
    \label{eq:moe2}
    G(x)=\mathrm{Softmax}(\mathrm{Top_k}(P(x),k))
\end{equation}
where $\mathrm{Top_k}$ function is given by \eqref{eq:moe3}:

\begin{equation}
    \label{eq:moe3}
    \mathrm{Top_k}(v_j, k)=\left\{
    \begin{aligned}
        & v_j, v_j~\mathrm{in}~\mathrm{top}~\mathrm{k}~\mathrm{elements}~\mathrm{of}~v \\
        & -\infty, \mathrm{otherwise}
    \end{aligned}
    \right.
\end{equation}

\subsection{Multi-Scale Contextualization}

In contrast to commonly used global contextualization, multi-scale contextualization is a local one. The nature of byte-based texts necessitates the involvement of local contextualization~\cite{tay2022charformer,sreedhar-etal-2023-local}. Then, MSC~\cite{huang-feng-2024-integrating} extends this to contextualization of various scales, adapting to different contexts.

The multi-scale contextualization functions given by MSC are simple and direct:
\begin{equation}
    g_i(\cdot, r)=\left\{
    \begin{aligned}
        & \mathrm{Identity}(\cdot) & , & ~k=0 \\
        & \mathrm{CNN}(\cdot,k) & , & ~k>0
    \end{aligned}
    \right.
    \label{eq:multiscale-contextualization}
\end{equation}
Where $k$ denotes the contextualization realm, i.e. the kernel size of CNN. Empirically, $k$ is chosen from {0, 1, 3, 5, 7}~\citep{huang-feng-2024-integrating}.

\subsection{Multi-Head Attention (MHA)}

\begin{figure*}[h]
    \centering
    \includegraphics[width=0.65\linewidth]{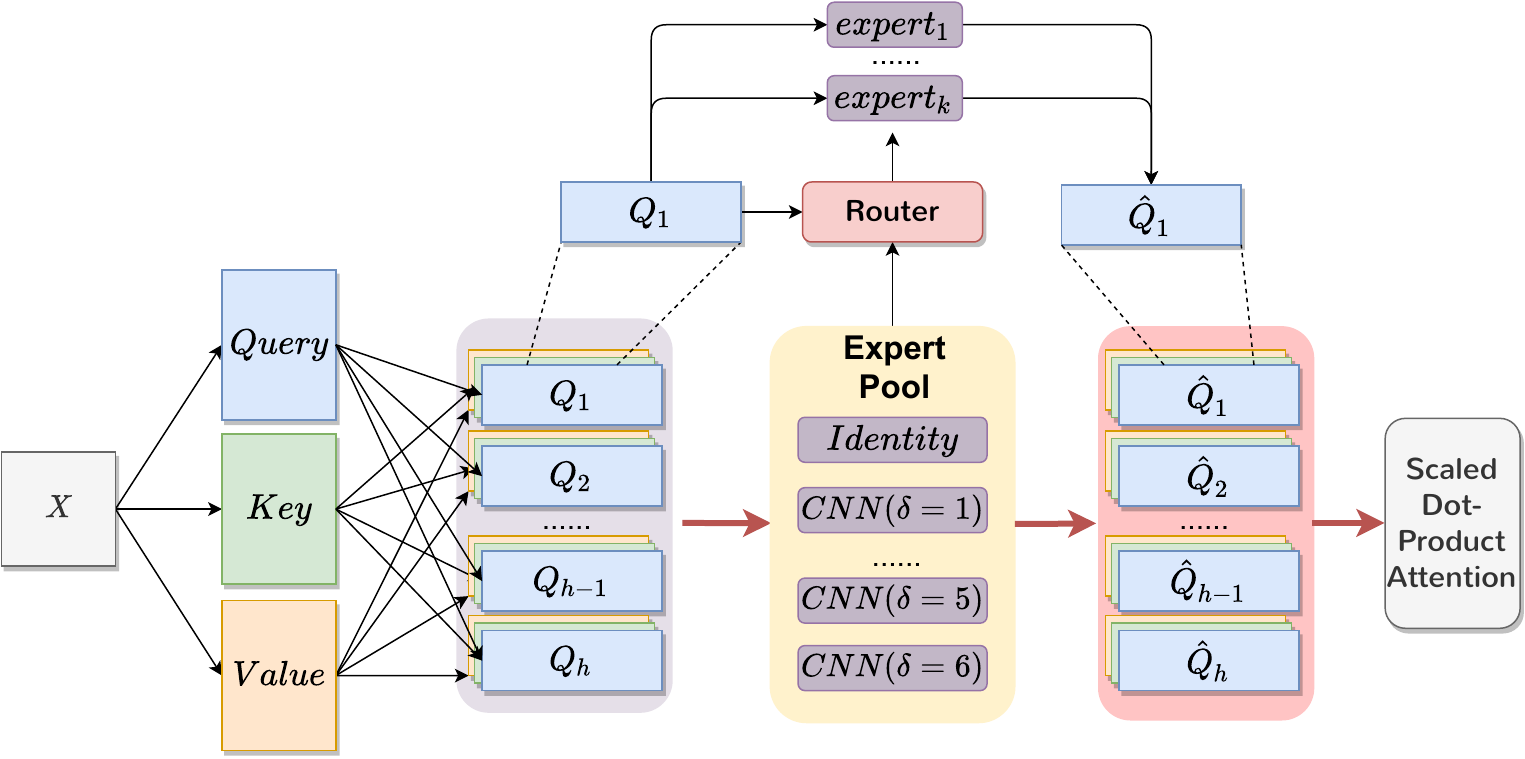}
    \caption{This figure shows the overall structure of MoCE. All contextualization functions are treated as experts. The router dynamically allocates experts from the expert pool with their corresponding weights for each head of each Q, K, and V vector. The heads are locally contextualized with given experts, and then serve as the input to the following Scaled Dot-Product Attention module.}
    \label{fig:model-overall}
\end{figure*}

MHA is one of the core components of Transformer structure~\cite{DBLP:conf/nips/VaswaniSPUJGKP17}, which employs multiple attention functions in parallel instead of a single one. Each attention function corresponds to an independent attention head, which linearly projects the input to Query, Key and Value vectors with a reduced dimension and applies "Scaled Dot-Product Attention"~\cite{DBLP:conf/nips/VaswaniSPUJGKP17} on them. Eventually, outputs of all heads are concatenated to a full dimension. 

An equivalent view of MHA is it breaks the hidden state dimensions of the linearly projected vectors into $h$ parts, as shown in \eqref{eq:split}, 
\begin{equation}
    \left\{
    \begin{array}{ccc}
        Q=&XW^Q=&[Q_1,Q_2,...,Q_h] \\
        K=&XW^K=&[K_1,K_2,...,K_h] \\
        V=&XW^V=&[V_1,V_2,...,V_h]
    \end{array}
    \right.
    \label{eq:split}
\end{equation}
and applies "Scaled Dot-Product Attention" for each, as shown in \eqref{eq:scaled-dot-product}. Finally, MHA is given by \eqref{eq:mha}.

\begin{equation}
\label{eq:scaled-dot-product}
    \mathrm{head}_i=\mathrm{softmax}(\frac{Q_iK_i^T}{\sqrt{d_k}})V_i
\end{equation}

\begin{equation}
\label{eq:mha}
    \mathrm{MHA}(X)=\mathrm{Cat}(\mathrm{head}_1,...,\mathrm{head}_h)W^O
\end{equation}

\subsection{Multilingual NMT}

For many-to-many multilingual NMT, it requires language IDs to indicate the source and target languages~\cite{johnson-etal-2017-googles}. A typical approach is prepending a language token at the beginning of the source and target sentences, respectively. Table \ref{language_token} provides an example.

\begin{table}[htbp]
\linespread{1.5}
\centering
 \resizebox{\linewidth}{!}{\begin{tabular}{c|cc}
\hline
 & Source & Target \\ \hline
Origin & Hello world! & Bonjour le monde! \\
Model Input  & <en> \_Hello \_world ! & <fr> \_Bonjour \_le \_monde ! \\ \hline
\end{tabular}}
\caption{An example of language token.}
\label{language_token}
\end{table}

\section{Method}
\begin{figure*}
    \centering
    \includegraphics[width=0.65\linewidth]{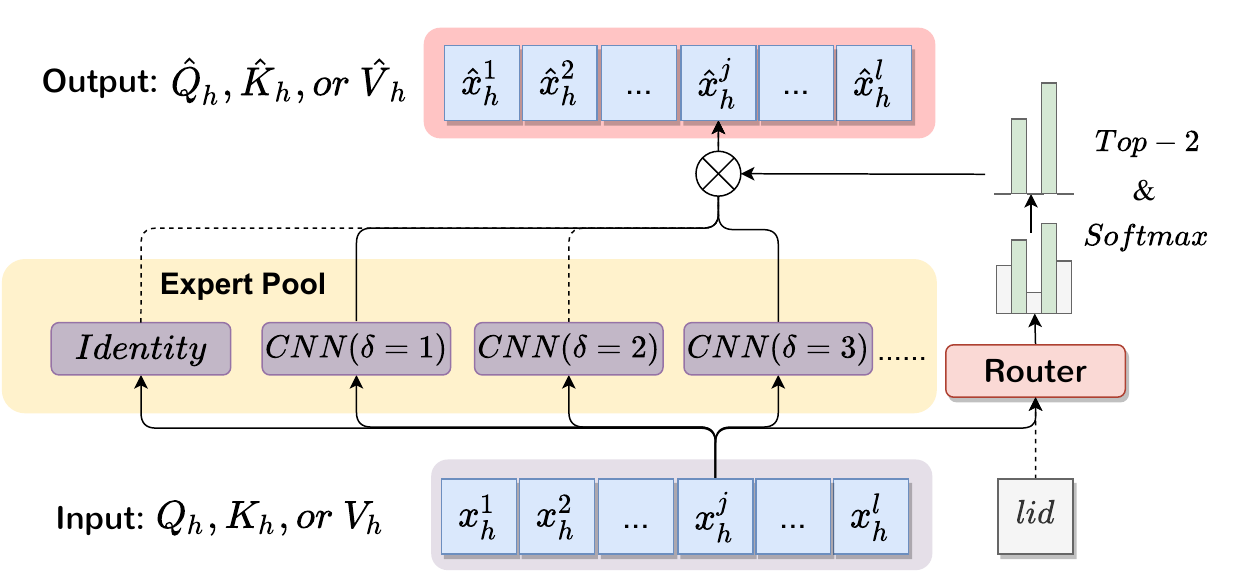}
    \caption{This figure shows the detailed model structure around the routing mechanism. The input $x_h$ denotes the $h^{th}$ head of Q, K, or V vector, with the sentence length of $l$. $x_h^j$ represents the $j^{th}$ token of $x_h$. For an arbitrary token $x_h^j$, the router predicts the selection probability for each expert and selects the top-2. The weighted combination of contextualized vectors from these 2 experts forms the final output $\hat{x}_h^j$. "$lid$" represents the language token, which is optional prior information for the router. If applied, "$lid$" is concatenated with $x_h^j$ to be the input.}
    \label{fig:model-router}
\end{figure*}

In this section, we first introduce MultiScale-Headed Attention, a way to model multi-scale contextualization. Then we extend it to Adaptive MultiScale-Headed Attention, the core module of MoCE. We also propose how to leverage language ID information to further improve our method.
% Leveraging multi-scale contextualization, MSC~\cite{huang-feng-2024-integrating} has achieved the state-of-the-art performance on byte-based NMT. However, it has a critical drawback that the contextualization scales are predefined and fixed. Worse still, it costs many trials to figure out a suitable set of these hyper-parameters, which can change largely among different datasets. In this section, we propose Mixture of Contextualization Experts (MoCE) that can adjust to not only different datasets but also various languages. In this section, we introduce the core module of MoCE, Ada-MSHA.

\subsection{MultiScale-Headed Attention (MSHA)}
Grouping hidden state dimensions has been a common and effective way to perform different operations on the same token~\cite{huang-feng-2024-integrating, Wu2024MultiHeadM}. Since it is the same as a part of MHA, we take advantage of MHA and use its heads as the grouped dimensions, like in \eqref{eq:split}. 

To perform multi-scale contextualization on each head, MSHA applies the same contextualization function $g(\cdot)$ on Q, K, and V vectors. Taking the $i^{th}$ head as an example, it substitutes the vectors in \eqref{eq:scaled-dot-product} with contextualized form, yielding \eqref{eq:ms-scaled-dot-product}.

\begin{equation}
\label{eq:ms-scaled-dot-product}
    \mathrm{\widetilde{head_i}}=\mathrm{softmax}(\frac{g_i(Q_i)g_i(K_i)^T}{\sqrt{d_k}})g_i(V_i)
\end{equation}

For $g(\cdot)$, we rewrite \eqref{eq:multiscale-contextualization} in a clear way as in \eqref{eq:multiscale-contextualization2}. Here $\delta$ denotes the neighborhood radius (including the central word) of local contextualization. 

\begin{equation}
    g(\cdot, \delta)=\left\{
    \begin{aligned}
        & \mathrm{Identity}(\cdot) & , & ~\delta=0 \\
        & \mathrm{CNN}(\cdot,2\delta-1) & , & ~\delta>0
    \end{aligned}
    \right.
    \label{eq:multiscale-contextualization2}
\end{equation}

Similarly, we replace $\mathrm{head}_i$ from \eqref{eq:ms-scaled-dot-product} with $\widetilde{\mathrm{head}_i}$. 

\begin{equation}
\label{eq:msha}
    \mathrm{MSHA}(X)=\mathrm{Cat}(\widetilde{\mathrm{head}_1},...,\widetilde{\mathrm{head}_h})W^O
\end{equation}

MSHA offers two advantages over MSC. First, it leverages the natural grouping operation within MHA, saving an additional vector separation and recombination step. Second, each head in MSHA is responsible for contextualization at a specific granularity, enhancing the model's interpretability, facilitating analysis presented in Section \ref{section:expert-ratio}.

\subsection{Adaptive MultiScale-Headed Attention}

Based on MSHA, we leverage MoE structure to propose an adaptive approach, Ada-MSHA, to solve the problem of fixed contextualization scales. 
 In our approach, contextualization functions $g(\cdot)$ are not determined by hyper-parameters. Rather, the model decides which $g(\cdot)$ to use. Specifically, a router predicts the selection probability of the candidate $g(\cdot)$ for each token respectively, and it selects $g(\cdot)$ according to the predicted probabilities. Viewing Ada-MSHA from the perspective of MoE, different contextualization functions serve as experts and they compose an expert pool, which the router selects experts from. The overall model structure is depicted in Figure \ref{fig:model-overall}.

As for routing mechanism, Ada-MSHA basically applies the commonly used implementation~\cite{shazeer2017,ramachandran2018diversity} for routing, i.e. \eqref{eq:moe1}, \eqref{eq:moe2}, and \eqref{eq:moe3}. 
% To align the notations, the $E_i(x)$ in \eqref{eq:moe1} is replaced with $g_i(x)$. 

Figure \ref{fig:model-router} shows the routing mechanism with $x_h^j$ as an example. The input $x$ can be any head of Q, K, or V vectors. $x_h^j$ means the $h^{th}$ head from the $j^{th}$ token of the sentence $x$. On one side, the router takes $x_h^j$ as input and predicts the selection probability for all experts. Then, the top-$k$, where $k$ is default as 2, probabilities are normalized to be a distribution. On the other side, the corresponding $k$ experts take $x_h^j$ as input and output the contextualized vector, $g_i(x_h^j)$. The final output $\hat{x}_h^j$ is given by the weighted summation of these $k$ vectors, as shown in \eqref{eq:our-moe}, which aligns with \eqref{eq:moe1}.

\begin{equation}
\label{eq:our-moe}
    \hat{x}_h^j=\sum_i^nG_i(x_h^j)g_i(x_h^j)
\end{equation}

The candidate contextualization functions, $g(\cdot)$, are manually defined in MSC~\cite{huang-feng-2024-integrating}, which has demonstrated it beneficial to provide $g(\cdot)$ of more scales. Therefore, we provide contextualization functions with various $\delta$s, consecutively from 0 to $\Delta$, the predefined upper bound. For example, if $\Delta=5$, there are 6 $g(\cdot)$, with $\delta=0,1,2,3,4,5$ for each. In this way, we reduce the number of hyper-parameters from 8 in MSC to only 1 in Ada-MSHA.

It should be clarified that Ada-MSHA is a modified Attention layer. By default, we replace the first encoder layer with Ada-MSHA and keep the rest model parts unchanged. This is discussed in Appendix \ref{appendix:position-of-adamsha-layer}.

\begin{figure*}[h]
    \centering
    \includegraphics[width=\linewidth]{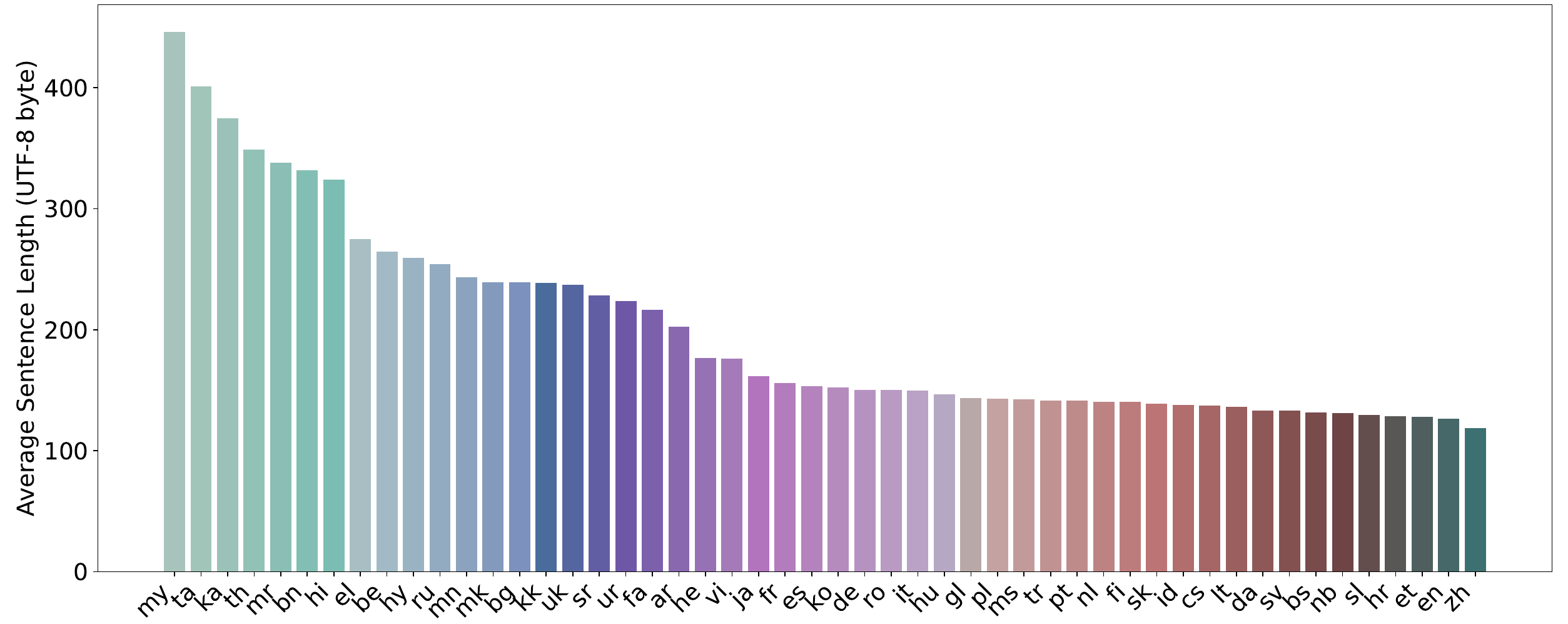}
    \caption{This figure depicts the average length of byte-based strings of different languages to express the same meaning. It reflects the complexity/conciseness of languages; languages are getting more concise from left to right.}
    \label{fig:avg-sentence-length}
\end{figure*}

\subsection{Language ID - the Prior Information}

The standard $P(x)$ from \eqref{eq:moe2} is given by \eqref{eq:router-msha}, where $W^R$ stands for the router and is a linear mapping from hidden state dimension to the number of expert candidates.
\begin{equation}
\label{eq:router-msha}
    P(x)=\mathrm{softmax}(xW^R)
\end{equation}

Realizing a byte may be interpreted differently as the language changes, we propose to concatenate the language ID (lid) token with $x$ to serve as router's input. The "$+lid$" version router is given by \eqref{eq:router-msha-lid}, where "$[\cdot|\cdot]$" denotes concatenation.

\begin{equation}
\label{eq:router-msha-lid}
    P(x)=\mathrm{softmax}([x|lid]W^R)
\end{equation}

The experiment results in \ref{section:effectiveness-lid} demonstrate the "$lid$" is beneficial to massively multilingual translation.

\section{Experiment Settings}
To verify the effectiveness of our proposed approach, we experiment on two massively multilingual translation datasets with over 50 languages. We mainly compare with byte-based models but also try the subword-based Transformer model.
\subsection{Datasets}

\subsubsection*{Ted-59}
Following \citet{huang-feng-2024-integrating}, we use Ted-59 dataset~\cite{qi-etal-2018-pre} for fair comparison. This dataset stems from the TED corpus and comprises 59 languages, including English and 58 other languages.
% whose data volume ranges from 3k to 210k. 
All sentence pairs are English-centered. We use the data provided by \citet{salesky-etal-2023-multilingual}.

For data preprocessing, we apply the scripts from EmbeddinglessNMT~\cite{shaham-levy-2021-neural} to perform byte-level preprocessing. The final vocabulary contains 256 bytes as well as several language tokens. For the subword-based system, SentencePiece~\cite{kudo-richardson-2018-sentencepiece} with vocabulary size 32k is applied for tokenization.

\subsubsection*{OPUS-100}
OPUS-100~\cite{zhang-etal-2020-improving} is an even larger massively multilingual translation dataset, which covers English and 99 other languages.

The data preprocessing is the same as that for Ted-59, except using Sentencepiece with vocabulary size 64k, aligning with \citet{zhang-etal-2020-improving}.

\subsection{Systems}
We compare our method with the following systems on multilingual benchmarks:

\begin{itemize}
    \item Transformer~\cite{DBLP:conf/nips/VaswaniSPUJGKP17}: Baseline that can be byte-based or subword-based.
    \item Byte-$n$CF~\cite{sreedhar-etal-2023-local}: A strong byte-based NMT model, with default hyper-parameter settings\footnote{\url{https://github.com/makeshn/LOBEF_Byte_NMT/blob/main/embeddingless_scripts/train_byte_ncf.sh}}.
    \item MSC~\cite{huang-feng-2024-integrating}: The state-of-the-art byte-based NMT model prior to Ada-MSHA. For Ted-59, we use recommended hyper-parameters\footnote{$k=0,0,3,3,5,5,7,7$ in \eqref{eq:multiscale-contextualization}}. For others, we use more contextualization scales which proves better\footnote{$k=0,0,1,1,3,5,5,7$ in \eqref{eq:multiscale-contextualization}}.
    \item MoCE: Our proposed method, with hyper-parameter $\Delta=5$ or $6$. Routing with language ID is also tested, shown as "$+lid$".
\end{itemize}

\begin{table}[h]
    \centering
    \begin{tabular}{lc}
        \hline
        Category    & Languages      \\ \hline
        Long        & my, ta, ka, th \\
        Medium         & bg, mk, uk, sr \\
        Short       & eo, sl, sv, et \\
        L.R.        & az, be, gl, sk \\
        H.R.        & ar, de, he, it \\
        OPUS4       & de, zh, br, te \\ \hline
    \end{tabular}
    \caption{Language composition for each category.}
    \label{table:language-selection}
\end{table}

\begin{table*}[h]
\centering
\begin{tabular}{l|c|ccc|cc|c}
\hline
         & Param. & Long  & Medium   & Short & L.R.  & H.R.  & All  \\ \hline
\begin{tabular}[c]{@{}c@{}}Transformer~(Subword)\\ \end{tabular} & 60.6M & 15.43 & 31.05 & 26.34 & 23.57 & 30.69 & 24.79 \\ 
\begin{tabular}[c]{@{}c@{}}Transformer~(Byte)\\ \end{tabular}    & 44.3M & 14.66 & 32.06 & 27.43 & 25.01 & 31.91 & 25.21 \\ \hline 
Byte-\textit{n}CF & 46.7M  & 13.75 & 31.09 & 26.31 & 23.55 & 30.84 & 24.33 \\ 
MSC     & 44.4M  & 14.86 & 32.36 & 28.00 & 25.11 & 32.26 & 25.61 \\ \hline 
MoCE~($\Delta=5$)     & 44.4M  & 15.89 & \textbf{33.23} & 28.56 & 25.81 & 32.92 & 26.30 \\ 
~~~~~~~~\textit{+lid} & 44.4M  & 16.28 & 33.19 & 28.64 & \textbf{25.84} & \textbf{33.02} & \textbf{26.52} \\ \hdashline
MoCE~($\Delta=6$)     & 44.5M  & 15.91 & 32.59 & 28.44 & 25.34 & 32.65 & 26.13 \\ 
~~~~~~~~\textit{+lid} & 44.5M  & \textbf{16.42} & 32.89 & \textbf{28.77} & 25.49 & \textbf{33.04} & 26.43 \\ \hline
\end{tabular}
\caption{Overall BLEU scores on Ted-59 dataset. The definition of each category is detailed in section \ref{section:language-selection}. "All" means the average score of all 58 translation directions.}
\label{table:main-ted59}
\end{table*}

\subsection{Training, Inference, and Evaluation}
We follow the standard practice in multilingual translation by training models on both "\textit{xx}$\rightarrow$\textit{en}" and "\textit{en}$\rightarrow$\textit{xx}" translation directions. Since the focus is handling diverse source languages, our evaluation primarily targets the "\textit{xx}$\rightarrow$\textit{en}" direction. Apart from reporting BLEU scores in the main body, we also report character-level metric ChrF and model-based metric COMET in Appendix \ref{appendix::comet} and \ref{appendix:chrf} for comprehensive evaluation. We detail the complete setups in Appendix \ref{appendix:settings}.

\section{Results and Analysis}
\label{section:results}
\subsection{Language Conciseness under UTF-8 Byte Encoding}
\label{section:language-conciseness}
Before translation experiments, we first examine the conciseness of different languages. As previously mentioned, a character may be represented by 1 to 4 UTF-8 bytes. A misleading intuition is that languages using 3-byte characters are longer in sentence lengths. However, languages like Chinese, are inherently concise, which is easily overlooked.

To explore language conciseness under byte encoding, we use Flores-101 dataset~\cite{goyal-etal-2022-flores}, which contains parallel sentence pairs across 101 languages, ensuring that all languages convey the same semantic content.
% The "devtest" subset includes 1012 parallel sentence pairs. 
Inspired by \citet{limisiewicz-etal-2024-myte}, we calculate the average length of byte-encoded sentences for each language, which reflects the conciseness of languages. The results are shown in Figure \ref{fig:avg-sentence-length}. The experiment details are introduced in Appendix \ref{appendix:language-conciseness}.
% According to the average sentence length, languages can be grouped into 3 categories: "Long", "Medium", and "Short". 

% \input{latex/Tables/language-selection}

\begin{table*}[h]
\centering
\begin{tabular}{l|c|ccc|c|c}
\hline
                      & Param. & Long           & Medium            & Short          & OPUS4          & All           \\ \hline
Transformer*~(Subword) & 110M   & -              & -              & -              & 23.35          & 27.60          \\ 
Transformer~(Subword)  & 77.0M  & 20.72          & 28.07          & 30.43          & 29.41          & 30.72          \\ 
Transformer~(Byte)     & 44.3M  & 16.63          & 24.47          & 24.84          & 22.81          & 25.36          \\ \hline
MSC                   & 44.4M  & 16.38          & 24.79          & 25.48          & 23.65          & 25.74          \\ \hline 
MoCE~($\Delta=5$)             & 44.4M  & 16.44          & 24.55          & 25.14          & 23.16          & 25.31          \\
~~~~~~~~\textit{+lid}                  & 44.4M  & 16.48          & \textbf{25.03} & 25.68          & 23.13          & 25.79          \\ \hdashline
MoCE~($\Delta=6$)             & 44.5M  & 16.48              & 24.92              & 25.47              & 23.58              & 25.79              \\
~~~~~~~~\textit{+lid}                  & 44.5M  & \textbf{17.06} & 24.94 & \textbf{25.93} & \textbf{24.01} & \textbf{26.10} \\ \hline
\end{tabular}
\caption{Overall BLEU scores on OPUS-100 dataset. 
% "Long", "Medium", "Short" and "All" follow the same definitions as in Table \ref{table:main-ted59}, and "OPUS4" denotes 4 languages selected by \citet{zhang-etal-2020-improving}, who proposed OPUS-100 dataset. The selected languages are presented in Appendix \ref{appendix:language-selection}. "*" sign shows the results from \citet{zhang-etal-2020-improving}. We bold the highest BLEU scores for byte-based methods.
The definition of each category is detailed in section \ref{section:language-selection}. "All" means the average score of all 94 translation directions. We bold the highest BLEU scores for byte-based methods.
}
\label{table:main-opus100}
\end{table*}

\subsection{Language Categorization}
\label{section:language-selection}
To effectively compare and analyze the results of multilingual translation, we report scores across different language categories. For each category, we select four representative languages and present their average scores. We first group languages into three categories: "Long", "Medium", and "Short", based on the average sentence length in section \ref{section:language-conciseness}. Following \citet{huang-feng-2024-integrating}, languages are grouped into low-resource (L.R.) and high-resource (H.R.) categories based on corpus size. Following \citet{zhang-etal-2020-improving}, four languages as selected as "OPUS4". The compositions of all categories are shown in Table \ref{table:language-selection}.

\subsection{Main Experiments}
\subsubsection*{Results on Ted-59}
Table \ref{table:main-ted59} summarizes the results on Ted-59 dataset. 
% The "Long", "Medium", and "Short" columns correspond to the language categories in section \ref{section:language-conciseness}. The "L.R" and "H.R" represent low-resource and high-resource language categories respectively, which follow \citet{huang-feng-2024-integrating}.
Nearly all byte-based models outperform the subword-based model with much fewer parameters. Among the byte-based models, all varieties of MoCE surpass MSC by a large margin, demonstrating the effectiveness of using adaptive contextualization scales.

The comparison of different settings of MoCE demonstrates two points. First, applying "$+lid$" achieves consistent improvement, which is discussed in Appendix \ref{section:effectiveness-lid}. Second, $\Delta$ value affects model's inclination toward language groups, especially "Long" and "Medium" groups, which is discussed in Section \ref{section:systematic-shift-delta}. The results at "Short" group do not follow a certain rule. We conjecture this is because a single byte has a determined mapping to character for "Short" languages, so they rely less on contextualization functions.

\subsubsection*{Results on OPUS-100}
Table \ref{table:main-opus100} exhibits the results on OPUS-100 dataset.\footnote{Byte-\textit{n}CF~\cite{sreedhar-etal-2023-local} is absent here because the source code fails to support such massive data volume, despite we used up our 300GB memory.} 
% The meanings of "Long", "Medium", and "Short" are the same as Ted-59. 
% "OPUS4" represents 4 languages selected by \citet{zhang-etal-2020-improving}. 
The "*" denotes the results from \citet{zhang-etal-2020-improving}, while our re-implementation is much better than the original one. 

The comparison between subword-based and byte-based models differs from Ted-59.
The reason lies in two sides.
First, the subword-based model doubles vocabulary size, reducing the risk of encountering unknown words. Second, OPUS-100 provides more training data, allowing effective convergence of the expanded embedding table.

The previous conclusion that $\Delta$ influences models' inclination toward different groups is less obvious also due to the sufficiency of training data. A larger $\Delta$ provides more expert choices, but a well-converged router can avoid choosing the inappropriate experts. As a result, $\Delta=6$ performs no worse than $\Delta=5$ in all groups.

\subsection{Expert Choice and Language Conciseness}
\label{section:expert-ratio}
The major motivation of MoCE is to choose appropriate contextualization functions according to the input. To verify if MoCE achieves such target, we count the selected ratio of each expert throughout a whole test set. Specifically, we select 4 languages, \textit{my}, \textit{bg}, \textit{et}, and \textit{zh}, from left to right in Figure \ref{fig:avg-sentence-length}, and also from least concise to most concise. Then, we test "\textit{en}$\rightarrow$\textit{xx}" and "\textit{xx}$\rightarrow$\textit{en}" for them, recording the selected ratios of experts. The sentence length ratios $xx/en$ are recorded to represent the relative conciseness, which highly correlates to the expert choices. All experiments are conducted on Ted-59 with "+$lid$" setting. 

\begin{figure*}[h]
    \centering
    \includegraphics[width=1\linewidth]{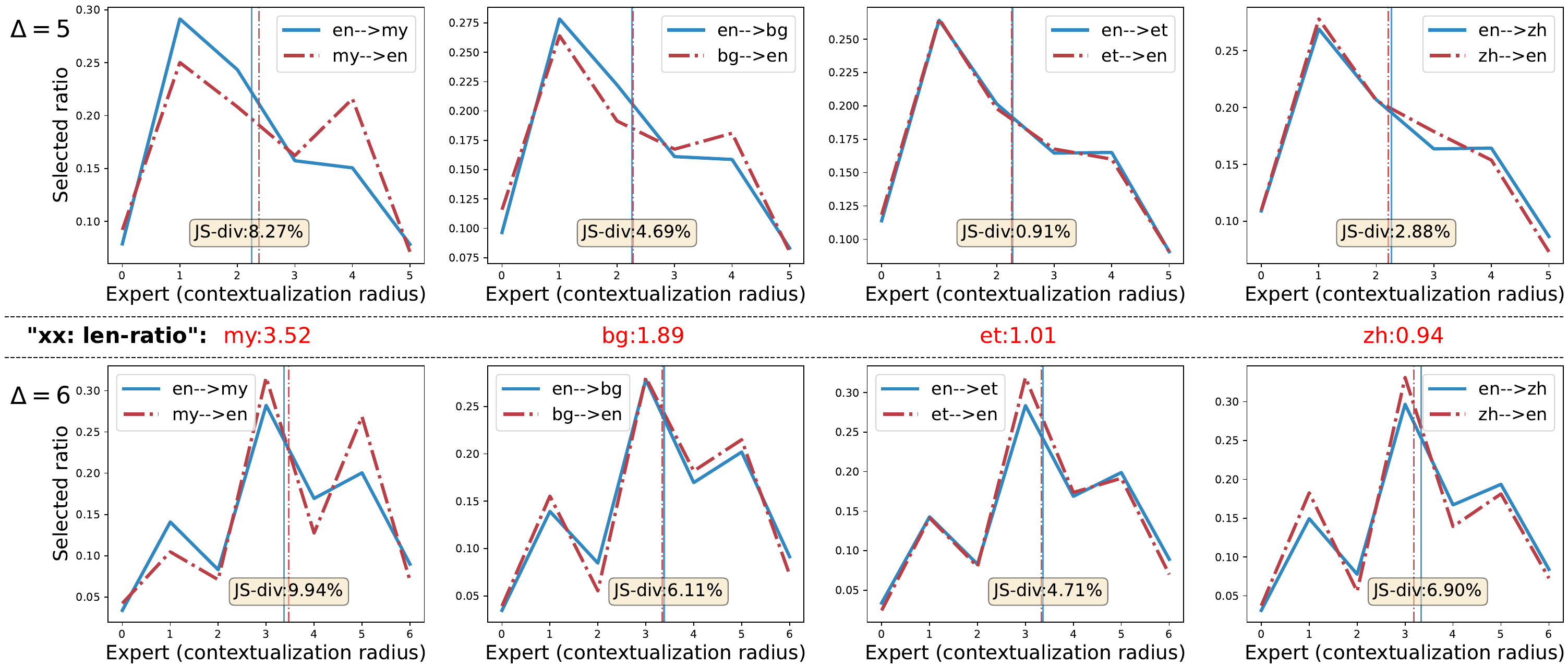}
    \caption{These figures exhibit the selected ratio of each expert, recorded when translating "\textit{en}$\rightarrow$\textit{xx}" and "\textit{xx}$\rightarrow$\textit{en}". The vertical lines are the averaged contextualization radius ($\delta$). The two rows represent $\Delta=5$ and $6$ respectively. Listed in the middle line are what "\textit{xx}" stands for and its length ratio over "\textit{en}", reflecting the relative conciseness. \textbf{From left to right}, as the language conciseness for "\textit{xx}" increases, the router tends to select more experts with smaller $\delta$. The JS-divergences between two distributions quantitatively prove it too. This consistency demonstrates that MoCE has learned to select proper experts for corresponding input. \textbf{From top to bottom}, as $\Delta$ increases, the model systematically tends to choose larger $\delta$. This explains why "$\Delta=6$" is better in "Long" and "$\Delta=5$" is better in "Medium" in Table \ref{table:main-ted59}.}
    \label{fig:exper-ratio}
\end{figure*}

The results are exhibited in Figure \ref{fig:exper-ratio}, where four columns represent different languages ("\textit{xx}"), and two rows represent model settings ($\Delta=5$ and $6$). The sentence length ratios of "\textit{xx}" over "\textit{en}" are listed in the middle line.
For all the sub-figures, the solid lines ("\textit{en}$\rightarrow$\textit{xx}") are treated as pivots, because they share the same source language. The differences between "\textit{xx}$\rightarrow$\textit{en}" and the pivots reveal model's inclination towards different languages. 

Taking either row as an example, the model gradually tends to choose smaller contextualization radius ($\delta$) as "\textit{xx}" becomes more concise. To quantitatively show this $\delta$-shift behavior, we drew the averages of $\delta$ in vertical lines. For the least concise language, "\textit{my}", model tends to choose larger $\delta$ than the pivot. For the most concise language, "\textit{zh}", model tends to choose smaller $\delta$ than the pivot. To measure the differences in more detail, we calculated the JS-divergence between the distributions of "\textit{xx}$\rightarrow$\textit{en}" and "\textit{en}$\rightarrow$\textit{xx}". The combination of figures and JS-divergence values clearly show the decreasing tendency of selected ratios for experts with larger $\delta$ and the increasing tendency of those with smaller $\delta$, from left to right. 

These findings demonstrate that MoCE can effectively identify the input language type and route the input to appropriate experts.

\subsection{Systematic Shift of $\delta$ with Varied $\Delta$}
\label{section:systematic-shift-delta}
Apart from the input language type, the choice of $\delta$ is also influenced by the upper bound of contextualization radius, which is $\Delta$. Intuitively, providing experts with larger $\delta$ encourages a model to leverage longer context information, though this is not always more beneficial for the overall performance. 

Comparing $\Delta=5$ and $6$ in Figure \ref{fig:exper-ratio}, we observe a consistent shift of $\delta$ selection. This shift of overall tendency is systematic, caused by model structure change. The vertical lines quantitatively demonstrate a model with larger $\Delta$ tends to select experts with a larger radius. This explains why $\Delta=6$ performs better in "Long" and $\Delta=5$ performs better in "Medium".

\subsection{How $\Delta$ Influences Translation Quality}
After analyzing how $\Delta$ influences expert choices, we empirically assess its influence on translation quality. Specifically, we conduct experiments on Ted-59 with different $\Delta$, and 
focus on different sentence length groups.
% test their performance on different sentence length groups. 
In addition to previous experiments, we try a smaller $\Delta=4$ for comparison. 

\begin{figure}[h!]
    \centering
    \includegraphics[width=0.9\linewidth]{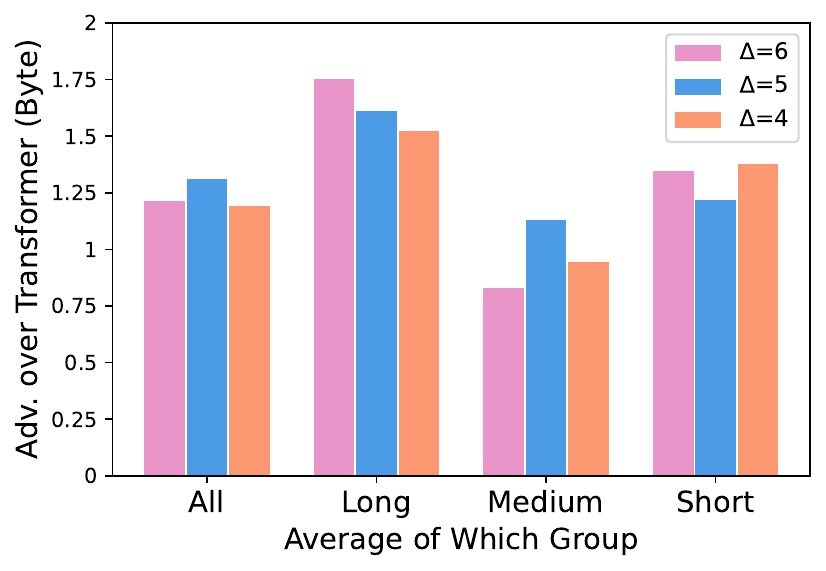}
    \caption{This figure compares MoCE with different $\Delta$ and show their advantage over Transformer (Byte). BLEU is used in this figure. Their performance gaps in different language groups show that $\Delta$ benefits languages with corresponding conciseness. 
    }
    \label{fig:analysis-scale}
\end{figure}

Figure \ref{fig:analysis-scale} exhibits their advantages over Transformer~(Byte). While $\Delta=6$ performs the best in "Long" group, $\Delta=5$ and $4$ perform the best in "Medium" group and "Short" group respectively. This figure demonstrates that for a certain language, using the appropriate contextualization function can maximize translation quality.

\subsection{Effectiveness of MoCE Concept}
We highlight the benefit of applying a Mixture of Contextualization Experts over using a single expert for each head. With the weights of experts varying in continuous space, the defacto contextualization radius has way more possibilities. Here, top-1 routing is chosen to simulate the single expert setting. Empirical results in Table \ref{table:top1} demonstrate the advantage of applying MoE in our method. It also proves that mixing two experts is enough.

\begin{table}[h]
\centering
\begin{tabular}{l|ccc|c}
\hline
\multicolumn{1}{c|}{} & Long  & Medium   & Short & Avg.  \\ \hline
Top-1                 & 15.86 & 32.74 & 28.29 & 26.03 \\ 
Top-2                 & \textbf{16.28} & \textbf{33.19} & \textbf{28.64} & \textbf{26.52} \\
Top-3                 & 16.15 & 32.90 & 28.51 & 26.34 \\ \hline
\end{tabular}
\caption{Top-2 routing is the standard MoCE, and Top-1 simulates the strategy of not mixing experts. Experiments are conducted with "$\Delta=5$, +$lid$" on Ted-59. }
\label{table:top1}
\end{table}

\subsection{Training and Inference Speed}
While MoCE demonstrates substantial improvements in translation quality over the baselines, the concern of time cost may rise. To address this concern, we evaluated the training and inference speed of all methods mentioned in this paper using Ted-59 dataset, providing a quantitative comparison of computational efficiency. The byte-based Transformer model serves as the baseline, the relative speed ratios are exhibited in Table \ref{table:speed}.

\begin{table}[h]
\centering
\begin{tabular}{|l|c|c|}
\hline
                    & Training & Inference \\ \hline
% Transformer(subword) & 1.5x  & 1.7x      \\ \hline
Transformer~(byte)    & 1x    & 1x        \\ \hline
MSC                 & 0.96x & 0.96x     \\ \hline
Byte-nCF            & 0.95x & 0.97x     \\ \hline
MoCE~($\Delta=5$)         & 0.98x & 0.96x     \\ \hline
~~~~~~~~+$lid$                & 0.99x & 0.96x     \\ \hline
MoCE~($\Delta=6$)         & 0.96x & 0.96x     \\ \hline
~~~~~~~~+$lid$                & 0.98x & 0.96x     \\ \hline
\end{tabular}

\caption{The relative speed ratio compared with byte-based Transformer baseline, measured on Ted-59 dataset.}
\label{table:speed}
\end{table}

For the training phase, we recorded the time spent in training one epoch. For the inference phase, we recorded the time spent in translating the entire test set for each translation direction, and took the sum of them. The speed of Transformer (byte) is denoted as "1x"; values below 1 indicate slower speeds compared to the baseline. According to Table \ref{table:speed}, all the local contextualization methods, including MoCE, achieve nearly the same speed as the baseline.

\section{Related Works}
\subsection{Byte-based Multilingual NMT}

Most multilingual NMT systems utilize subword-level tokenization techniques such as Byte Pair Encoding~\cite{sennrich-etal-2016-neural} and Unigram~\cite{kudo-2018-subword}. They often result in over-segmentation of low-resource languages, issues with out-of-vocabulary (OOV) words, and difficulties in adapting to new languages or domains~\cite{rust-etal-2021-good, raunak-etal-2020-long}. As the number of languages increases, these problems worsen, limiting the performance of multilingual NMT systems. To this end, 
various byte-level approaches have been proposed.
\citet{shaham-levy-2021-neural} build embedding-free NMT models using UTF-8 byte tokenization.
\citet{tay2022charformer} propose CharFormer,
integrating
byte representations from different block sizes by mean-pooling. \citet{sreedhar-etal-2023-local} enhance CharFormer by employing
CNN
for representation integration. \citet{huang-feng-2024-integrating} propose 
MSC,
which assigns contextualized information at different scales across various hidden state dimensions.

\subsection{MoE for Multilingual NMT}

Previous studies on multilingual NMT have demonstrated that model performance initially improves but subsequently declines as the number of training languages and the corpus size increase~\cite{DBLP:journals/corr/abs-1907-05019, zhang-etal-2020-improving, yang-etal-2021-improving-multilingual}. This phenomenon, known as the "curse of multilingualism", is attributed to limited model capacity~\cite{chang-etal-2023-multilinguality}. The MoE techniques significantly scale up model capacity without increasing training and inference costs accordingly, and can be used to scale up multilingual NMT models and mitigate language conflicts.
\citet{shazeer2017} propose sparse-gated modules with thousands of experts and validate the effect on bilingual translation tasks. \citet{li-etal-2023-mmnmt} use pre-trained FFN to initialize each expert, and the model can flexibly combine dense and sparse MoE modules. To address the problem of insufficient expert training in sparse MoE, \citet{Wu2024MultiHeadM} integrates the knowledge of experts through token-splitting-merging operation, which significantly improves the expert activation ratio.

\subsection{Local Contextualization Modeling}
% While the global contextualization feature of the attention mechanism addressed the long-distance dependency problem, many tasks require local contextualization. For example, it takes a model a few consecutive frames to understand a piece of audio. Conformer~\cite{DBLP:conf/interspeech/GulatiQCPZYHWZW20} has proof to be an effective local contextualization structure in audio-related tasks, such as speech recognition~\cite{DBLP:conf/interspeech/GulatiQCPZYHWZW20}, speech translation~\cite{fang-etal-2024-ctc}, and simultaneous translation~\cite{ma-etal-2024-non}. Since Ada-MSHA achieves similar function as Conformer, it may be a better replacement for Conformer. Other tasks, like modeling phrasal representation~\cite{huang-etal-2023-enhancing, fang-feng-2022-neural}, can also leverage Ada-MSHA as a good multi-scale feature extractor.

While the attention mechanism’s global contextualization effectively addresses the challenge of long-distance dependencies, many tasks benefit from a focus on local contextualization. For instance, in audio processing, a model needs several consecutive frames to comprehend an audio segment. The Conformer~\cite{DBLP:conf/interspeech/GulatiQCPZYHWZW20} has proven to be an effective local contextualization structure for audio-related tasks such as speech recognition~\cite{DBLP:conf/interspeech/GulatiQCPZYHWZW20}, speech translation~\cite{fang-etal-2024-ctc}, and simultaneous translation~\cite{ma-etal-2024-non}. Given that MoCE provides similar functionality to Conformer, it may serve as a more advantageous replacement. Moreover, tasks involving phrasal representation modeling~\cite{huang-etal-2023-enhancing, fang-feng-2022-neural} could also benefit from MoCE’s capacity as a robust multi-scale feature extractor.

\section{Conclusions}
In this work, we propose an Adaptive MultiScale-Headed Attention approach for effective local contextualization. It conducts contextualization on attention heads within multi-head attention. Next, we leverage the concept of MoE to achieve adaptive selection of contextualization functions. Extensive experiments demonstrate the superiority of Ada-MSHA over other byte-based models and disclose why it performs well.

\section*{Limitations}
The proposed approach is only applied to encoder. How to apply local contextualization to decoder is a critical and interesting topic for future work.

% \section*{Acknowledgments}

% \bibliography{custom}
\bibliography{custom,anthology}

\newpage
\onecolumn
\appendix
\section{Language Conciseness}
\label{appendix:language-conciseness}
The Flores-100 dataset contains two subset, "dev" and "devtest". Here, we use "devtest" subset which includes 1012 parallel sentence pairs.

We first convert all sentences into byte-based strings, and then count the average sentence length for each language. Since the sentences are parallel across all languages, the average sentence lengths are comparable, showing how many bytes the language need to convey the same meaning.

Figure \ref{fig:avg-sentence-length} shows a subset of languages from Flores-100. In fact, we choose the intersection set of Flores-100 and Ted-59.
\label{appendix:sentence-length}
% \begin{figure*}[h]
%     \centering
%     \includegraphics[width=\linewidth]{latex/Figs/lang-avg-len.pdf}
%     \caption{This figure depicts the average length of byte-based string of the languages, which reflects the language conciseness.}
%     \label{fig:avg-sentence-length}
% \end{figure*}

% \section{Selection of Reported Languages}
% \label{appendix:language-selection}
% According to sentence lengths, as presented in Figure \ref{fig:avg-sentence-length}, we select 4 representative languages for "Long", "Medium", and "Short" groups for testing. From the perspective of corpus size, we select 4 low-resource (L.R.) languages and four high-resource (H.R.) languages for testing. We also test on 4 languages selected by \citet{zhang-etal-2020-improving}, which is denoted as "OPUS4". The language selection is shown in Table \ref{table:language-selection}.

% \input{latex/Tables/language-selection}

\section{Detailed Training, Inference and Evaluation Setup}
\label{appendix:settings}
All the models and experiments are implemented based on Fairseq~\cite{ott-etal-2019-fairseq} codebase.

During training stage, we set learning rate=$5e-4$, dropout=$0.1$, and label smoothing=$0.1$. The batch size is 65536 for Ted-59 and PC-6 datasets, and 131072 for the largest OPUS-100 dataset. We apply adam optimizer with $\beta=(0.9, 0.98)$ and $\epsilon=1e-8$. In our experiments, encoder and decoder always share the embeddings. We also apply an early-stop strategy, i.e. stop training until valid loss doesn't decrease for 10 checkpoints. Checkpoints are saved for every 5k update steps.

During inference, we average the last 5 checkpoints and evaluate with it. We apply beam search with beam size=$4$. We observed that byte-based NMT models tend to generate longer sequences, so we set the length penalty=$1.5$ for all byte-based models. Note that OPUS-100 lacks validation set and test set for these 5 languages: Aragonese(an), Divehi(dz), Yoruba(yo), Mongolian(mn), and Armenian(hy). In our experiments, these 5 languages are still trained, but we do not test them. 

For comprehensive evaluation, we report BLEU~\cite{PapineniRWZ02} and ChrF~\cite{popovic-2015-chrf} scores using \textit{SacreBLEU} toolkit~\cite{post-2018-call}, and report COMET~\cite{rei-etal-2022-comet} score using \textit{wmt22-comet-da}\footnote{\url{https://huggingface.co/Unbabel/wmt22-comet-da}} model. To keep the article concise, we present BLEU scores in the main body, and present ChrF and COMET in the Appendix \ref{appendix:chrf} and \ref{appendix::comet}, respectively.

\section{Applying MoCE on Subword-based Models}
The experiment results in Section \ref{section:results} have demonstrated the significant effectiveness of MoCE on byte-based models. In addition to that, we also explored how it benefits subword-based models. Intuitively, subword tokens contain richer information and require less contextual information. Therefore, we experimented on shorter contextual scales, i.e. $\Delta=4$ and $\Delta=5$. We report the BLEU scores on Table \ref{table:ted59-subword}. According to the results, applying MoCE improves subword-based models, though the improvement is relatively minor. We conjecture this is because the subword-based vocabulary still suffers from the curse of multilinguality.
\begin{table}[h]
\centering
\begin{tabular}{l|ccc|cc|c}
\hline
                             & Long  & Medium & Short & L.R.  & H.R.  & All   \\ \hline
Transformer                  & 15.43 & 31.05  & 26.34 & 23.57 & 30.69 & 24.79 \\ \hline
MoCE~($\Delta=4$)               & \textbf{15.71} & 31.15  & 26.65 & 23.49 & 31.05 & 24.99 \\
~~~~~~~~\textit{+lid} & 15.51 & \textbf{31.46}  & 26.66 & 23.77 & 31.10 & 24.98 \\ \hdashline
MoCE~($\Delta=5$)               & 15.49 & 30.98  & 26.93 & 24.10 & 31.09 & 25.01 \\
~~~~~~~~\textit{+lid} & 15.63 & \textbf{31.46}  & \textbf{27.26} & \textbf{24.26} & \textbf{31.59} & \textbf{25.34} \\ \hline
\end{tabular}
\caption{BLEU scores on Ted-59 dataset. All models are subword-based.}
\label{table:ted59-subword}
\end{table}

\section{ChrF Scores on Ted-59 and OPUS-100}
\label{appendix:chrf}
We report ChrF scores on Ted-59 and OPUS-100 test sets in Table \ref{table:chrf2-ted59} and \ref{table:chrf2-opus100}, respectively.
\begin{table*}[h]
\centering
\begin{tabular}{l|c|ccc|cc|c}
\hline
         & Param. & Long  & Medium   & Short & L.R.  & H.R.  & All  \\ \hline
\begin{tabular}[c]{@{}c@{}}Transformer~(Subword)\\ \end{tabular} & 60.6M & 37.44 & 53.05 & 48.15 & 45.74 & 52.41 & 46.70 \\ 
\begin{tabular}[c]{@{}c@{}}Transformer~(Byte)\\ \end{tabular}    & 44.3M & 36.82 & 54.30 & 49.42 & 47.55 & 53.45 & 47.26 \\ \hline 
Byte-\textit{n}CF & 46.7M  & 35.36 & 53.02 & 48.27 & 46.03 & 52.17 & 46.09 \\ 
MSC     & 44.4M  & 36.88 & 54.53 & 49.86 & 47.69 & 53.64 & 47.54 \\ \hline 
MoCE~($\Delta=5$)     & 44.4M  & 38.47 & 55.07 & 50.60 & 48.19 & 54.27 & 48.30 \\ 
~~~~~~~~\textit{+lid} & 44.4M  & 38.95 & \textbf{55.14} & 50.57 & \textbf{48.23} & \textbf{54.45} & \textbf{48.56} \\ \hdashline
MoCE~($\Delta=6$)     & 44.5M  & 38.98 & 54.80 & 50.54 & 47.78 & 54.29 & 48.34 \\ 
~~~~~~~~\textit{+lid} & 44.5M  & \textbf{39.04} & 54.77 & \textbf{50.66} & 47.91 & 54.37 & 48.42 \\ \hline
\end{tabular}
\caption{Overall chrF scores on Ted-59 dataset. "Long", "Medium", "Short", "L.R.", "H.R." and "All" follow the same definitions as in Table \ref{table:main-ted59}.}
\label{table:chrf2-ted59}
\end{table*}

\begin{table*}[h]
\centering
\begin{tabular}{l|c|ccc|c|c}
\hline
                      & Param. & Long           & Medium            & Short          & OPUS4          & All           \\ \hline
Transformer~(Subword)  & 77.0M  & 39.97 & 48.02 & 50.36 & 49.13 & 49.96          \\ 
Transformer~(Byte)     & 44.3M  & 35.37 & 44.85 & 44.93 & 42.25 & 44.88          \\ \hline
MSC                   & 44.4M  & 35.39 & \textbf{45.30} & 45.54 & 43.24 & 45.11          \\ \hline 
MoCE~($\Delta=5$)      & 44.4M  & 34.84 & 44.84 & 45.17 & 42.69 & 44.62          \\
~~~~~~~~\textit{+lid} & 44.4M  & 35.27 & 45.26 & 45.59 & 42.74 & 45.13          \\ \hdashline
% ~~~~~~~~\textit{+lid} & 44.4M  & 35.36 & 45.26 & 45.63 & 42.79 & 45.10          \\ \hdashline
MoCE~($\Delta=6$)      & 44.5M  & 35.01 & 45.08 & 45.49 & 43.01 & 45.14              \\
~~~~~~~~\textit{+lid} & 44.5M  & \textbf{35.77} & 45.27 & \textbf{45.86} & \textbf{43.52} & \textbf{45.37} \\ \hline
% ~~~~~~~~\textit{+lid} & 44.5M  & \textbf{35.53} & 45.24 & \textbf{45.84} & \textbf{43.50} & \textbf{45.27} \\ \hline
\end{tabular}
\caption{Overall chrF scores on OPUS-100 dataset. "Long", "Medium", "Short", "OPUS4" and "All" follow the same definitions as in Table \ref{table:main-opus100}.}
\label{table:chrf2-opus100}
\end{table*}

% \newpage

\section{Comet Scores on Ted-59 and OPUS-100}
\label{appendix::comet}
We report COMET scores on Ted-59 and OPUS-100 test sets in Table \ref{table:comet-ted59} and \ref{table:comet-opus100}, respectively.

\begin{table*}[h!]
\centering
\begin{tabular}{l|c|ccc|cc|c}
\hline
         & Param. & Long  & Medium   & Short & L.R.  & H.R.  & All  \\ \hline
\begin{tabular}[c]{@{}c@{}}Transformer~(Subword)\\ \end{tabular} & 60.6M & 69.41 & 77.64 & 74.49 & 73.58 & 77.39 & 74.46 \\ 
\begin{tabular}[c]{@{}c@{}}Transformer~(Byte)\\ \end{tabular}    & 44.3M & 67.96 & 79.11 & 74.87 & 74.88 & 77.85 & 74.44 \\ \hline 
Byte-\textit{n}CF & 46.7M  & 66.08 & 77.30 & 73.23 & 72.66 & 76.29 & 72.84 \\ 
MSC     & 44.4M  & 68.11 & 79.24 & 75.54 & 75.17 & 78.20 & 74.81 \\ \hline 
MoCE~($\Delta=5$)     & 44.4M  & 69.74 & \textbf{79.89} & 76.40 & 75.93 & 79.03 & 75.79 \\ 
~~~~~~~~\textit{+lid} & 44.4M  & 70.40 & 79.88 & 76.53 & \textbf{76.05} & \textbf{79.34} & \textbf{76.12} \\ \hdashline
MoCE~($\Delta=6$)     & 44.5M  & \textbf{70.65} & 79.69 & 76.58 & 75.85 & 79.22 & 76.09 \\ 
~~~~~~~~\textit{+lid} & 44.5M  & 70.58 & 79.71 & \textbf{76.73} & 75.77 & 79.29 & 76.10 \\ \hline
\end{tabular}
\caption{Overall COMET scores on Ted-59 dataset. "Long", "Medium", "Short", "L.R.", "H.R." and "All" follow the same definitions as in Table \ref{table:main-ted59}.}
\label{table:comet-ted59}
\end{table*}

\begin{table*}[h!]
\centering
\begin{tabular}{l|c|ccc|c|c}
\hline
                      & Param. & Long           & Medium            & Short          & OPUS4          & All           \\ \hline
Transformer~(Subword)  & 77.0M  & 74.42 & 76.22 & 77.53 & 75.86 & 77.00          \\ 
Transformer~(Byte)     & 44.3M  & 71.14 & 74.11 & 74.62 & 72.41 & 74.64          \\ \hline
MSC                   & 44.4M  & 71.21 & 74.42 & 75.34 & 73.12 & 74.69          \\ \hline 
MoCE~($\Delta=5$)      & 44.4M  & 70.44 & 73.84 & 74.76 & 72.42 & 74.13          \\
~~~~~~~~\textit{+lid} & 44.4M  & 71.06 & 74.41 & 75.18 & 72.88 & 74.82          \\ \hdashline
% ~~~~~~~~\textit{+lid} & 44.4M  & 71.00 & 74.37 & 75.23 & 72.88 & 74.78          \\ \hdashline
MoCE~($\Delta=6$)      & 44.5M  & 70.75 & 74.28 & 75.05 & 72.81 & 74.81              \\
~~~~~~~~\textit{+lid} & 44.5M  & \textbf{71.32} & \textbf{74.51} & \textbf{75.50} & \textbf{73.27} & \textbf{74.89} \\ \hline
% ~~~~~~~~\textit{+lid} & 44.5M  & \textbf{71.26} & \textbf{74.47} & \textbf{75.46} & \textbf{73.18} & \textbf{74.87} \\ \hline
\end{tabular}
\caption{Overall COMET scores on OPUS-100 dataset. "Long", "Medium", "Short", "OPUS4" and "All" follow the same definitions as in Table \ref{table:main-opus100}.}
\label{table:comet-opus100}
\end{table*}

\section{Experiment on a Small Multilingual Translation Dataset}
\label{appendix:pc6}
PC-6~\cite{bu-etal-2024-improving} is a small multilingual translation dataset with English and 5 other languages. The purpose of using PC-6 is mainly to investigate the effectiveness of "$lid$" in few-language scenario. Therefore, we only conduct byte-level preprocess, which is the same as that for Ted-59. For the same reason, we simply train and test models on solely "xx$\rightarrow$en" direction.

Table \ref{table:pc6} shows the experiment results on PC-6 dataset. We find that "$+lid$" does not work on this dataset. We conjecture this is because when language numbers are limited, the router already knows how to select experts and does not rely on the prior information.
\begin{table*}[h!]
\centering
\begin{tabular}{l|c|ccccc|c}
\hline
\multicolumn{1}{c|}{} & Param. & cs    & kk    & ro    & ru    & tr             & All  \\ \hline
Transformer~(Byte)     & 44.3M  & 21.25 & 9.78  & 29.67 & 27.59 & 18.53          & 21.36 \\ \hline
Byte-nCF              & 46.6M  & 20.65 & 9.30  & 29.01 & 27.19 & 18.36          & 20.91 \\
MSC                   & 45.0M  & 19.44 & 9.19  & 28.90 & 27.48 & 18.26          & 20.66 \\ \hline
MoCE~($\Delta=5$) & 44.4M & \textbf{21.84} & \textbf{10.99} & \textbf{29.95} & 28.22          & 19.17 & \textbf{22.03} \\
~~~~~~~~\textit{+lid}                   & 44.4M  & 21.48 & 10.64 & 29.55 & 27.38 & \textbf{19.22} & 21.65 \\ \hdashline
MoCE~($\Delta=6$) & 44.5M & 21.09          & 10.66          & 29.76          & \textbf{28.34} & 18.60 & 21.69          \\
~~~~~~~~\textit{+lid}                   & 44.5M  & 21.72 & 10.81 & 29.61 & 28.21 & 18.66          & 21.80 \\ \hline
\end{tabular}
\caption{Overall BLEU scores on PC-6 dataset."All" follows the same definition as in Table \ref{table:main-ted59}. We bold the highest BLEU scores.}
\label{table:pc6}
\end{table*}

\section{The Best Position to Place Ada-MSHA Layer}
\label{appendix:position-of-adamsha-layer}
\citet{huang-feng-2024-integrating} has mentioned the best position to place MSC layer is the first one. Here, we conduct an experiment on Ted-59 dataset to explore the best position for Ada-MSHA layer. 

\begin{figure}[h]
    \centering
    \includegraphics[width=0.6\linewidth]{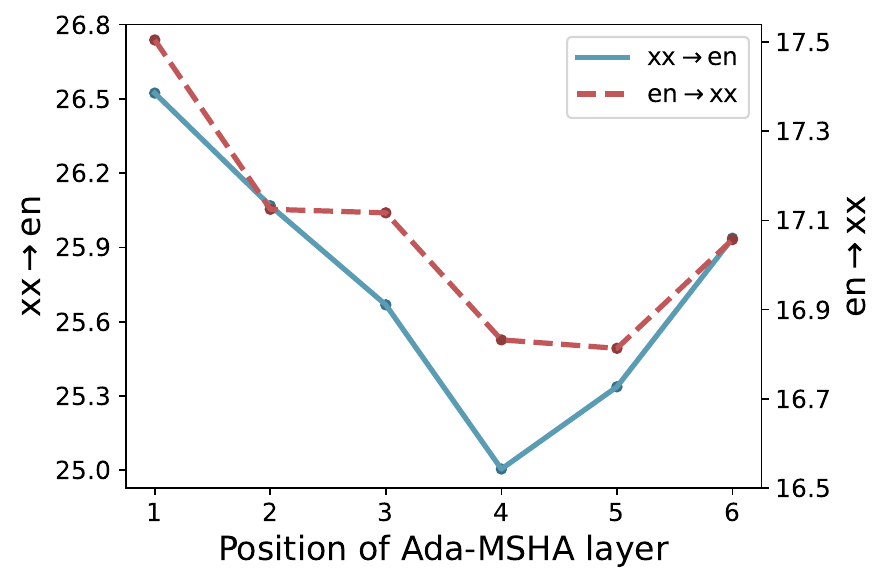}
    \caption{An empirical study on the best position to place Ada-MSHA layer. According to this figure, the first layer should be the best position, which aligns with the conclusion in MSC~\cite{huang-feng-2024-integrating}.}
    \label{fig:analyse-layer-position}
\end{figure}

The results in Figure \ref{fig:analyse-layer-position} shows that placing Ada-MSHA at the first layer is the best choice. It is not hard to figure out the explanation to this phenomenon. When placing Ada-MSHA at the first layer, the local contextualization can be treated as a part of the embedding layer, which resembles ELMo~\cite{peters-etal-2018-deep}. At the latter layers, however, the vectors have already been globally contextualized by attention, and conducting local contextualization now makes little sense.

% \section{Speed Evaluation of different models}
% While Ada-MSHA demonstrates substantial improvements in translation quality over the baselines, the concern of time cost may rise. To address this concern, we evaluated the training and inference speeds of all methods mentioned in this paper using Ted-59 dataset, providing a quantitative comparison of computational efficiency. The byte-based Transformer model serves as the baseline, the relative speed ratios are exhibited.

% Here is how we measure the speeds of both phases. For the training phase, we recorded the time spent in training one epoch. For the inference phase, we recorded the time spent in translating the entire test set for each translation direction, and took the sum of them. The comprehensive results are presented in Table \ref{table:speed}, where the speed of Transformer(byte) is denoted as "1x"; values below 1 indicate slower speeds compared to the baseline. As shown in this table, all the local contextualization methods, including Ada-MSHA, achieve nearly the same speed as the baseline.

% \input{latex/Tables/speeds}

\section{Experiment Under Low Resource Setting}
Intuitively, byte-based text patterns are more difficult for a model to learn, because the free combination of bytes results in a much larger language model search space. This problem should be worse under low-resource settings, where less data are utilized to constrain the model. We hypothesize that this can be mitigated by properly contextualizing, which helps models focus on local patterns. To better understand how different models perform under low-resource settings, we curated a low-resource dataset from Ted-59.

Specifically, we selected all languages whose training dataset has less than 50k parallel sentences and conducted experiments on this sub-dataset. We further split these languages into two categories: low-resource (L.R.), which has more than 10k parallel sentences, and extremely low-resource (E.L.R.), which has less than 10k parallel sentences. The corresponding languages are listed in Table \ref{table:lr-language-selection}. 

The training, inference, and evaluation details are almost the same as those for Ted-59 dataset. The only difference is the representative languages for the Long, Medium, and Short categories. Since some of the languages from Table \ref{table:language-selection} are not low resources, we re-selected four languages for each category, the first two from L.R. and the last two from E.L.R., according to the order in Figure \ref{fig:avg-sentence-length}. For MoCE, we experiment only with the best model setting, $\Delta=5$ with "lid".

The experiment results are exhibited in Table \ref{table:results-low-resource}. Surprisingly, byte-based Transformer fails the tasks, while subword-based Transformer performs adequately. This demonstrates the difficulty of learning byte-based text with only global attention. Compared with byte-based Transformer, other local contextualization-based methods are able to learn the patterns with the low-resourced training data. Among them, MoCE shows significant improvement over other methods, and the advantages are larger than that on the full Ted-59 dataset. This demonstrates our proposed method, which adaptively chooses contextualization neighbors, requires fewer data to converge to an optimal solution.

\begin{table}[h]
\centering
\begin{tabular}{l|ccc|cc|c}
\hline
                     & Long  & Medium & Short & L.R.    & E.L.R.   & All   \\ \hline
Transformer~(Subword) & 10.89 & 17.20  & 18.53 & 23.09 & 12.99 & 18.56 \\
Transformer~(Byte)    & 1.01  & 1.14   & 0.86  & 0.96  & 1.17  & 1.05  \\ \hline
Byte-nCF             & 10.14 & 16.58  & 17.39 & 22.45 & 12.10 & 17.81 \\
MSC                  & 10.14 & 15.80  & 17.01 & 22.16 & 11.76 & 17.50 \\ \hline
MoCE~($\Delta=5$) & 11.31 & 17.65  & 18.61 & \textbf{24.03} & 13.45 & 19.10 \\ 
~~~~~~~~+$lid$ & \textbf{11.80} & \textbf{18.21}  & 18.45 & \textbf{24.03} & \textbf{13.81} & \textbf{19.45} \\ \hline
\end{tabular}
\caption{The BLEU scores on the low-resource subset of Ted-59. In this table, "L.R." is the abbreviation of "Low Resource", meaning languages with training data size less than 50k but larger than 10k. "E.L.R" is the abbreviation of "Extremely Low Resource", meaning languages with training data size less than 10k. For each category, we average the scores of all languages.}
\label{table:results-low-resource}
\end{table}
\begin{table*}[h]
    \centering
    \begin{tabular}{lc}
        \hline
        Category    & Languages      \\ \hline
        L.R. & et, ku, nb, sl, hy, lt, ka, hi, \\
         & mk, sq, sk, gl, my, da, fi, frca \\
        E.L.R. & bs, ur, zh, ms, bn, ta, be \\
        & eu, mr, kk, eo, mn, az \\ \hline
        Long        & my, ka, ta, mr \\
        Medium      & mk, fi, kk, ms \\
        Short       & et, sl, zh, bs \\ \hline
    \end{tabular}
    \caption{Language selection for each category.}
    \label{table:lr-language-selection}
\end{table*}

\section{Effectiveness of Using Language ID}
\label{section:effectiveness-lid}
\begin{figure}[h]
    \centering
    \includegraphics[width=0.5\linewidth]{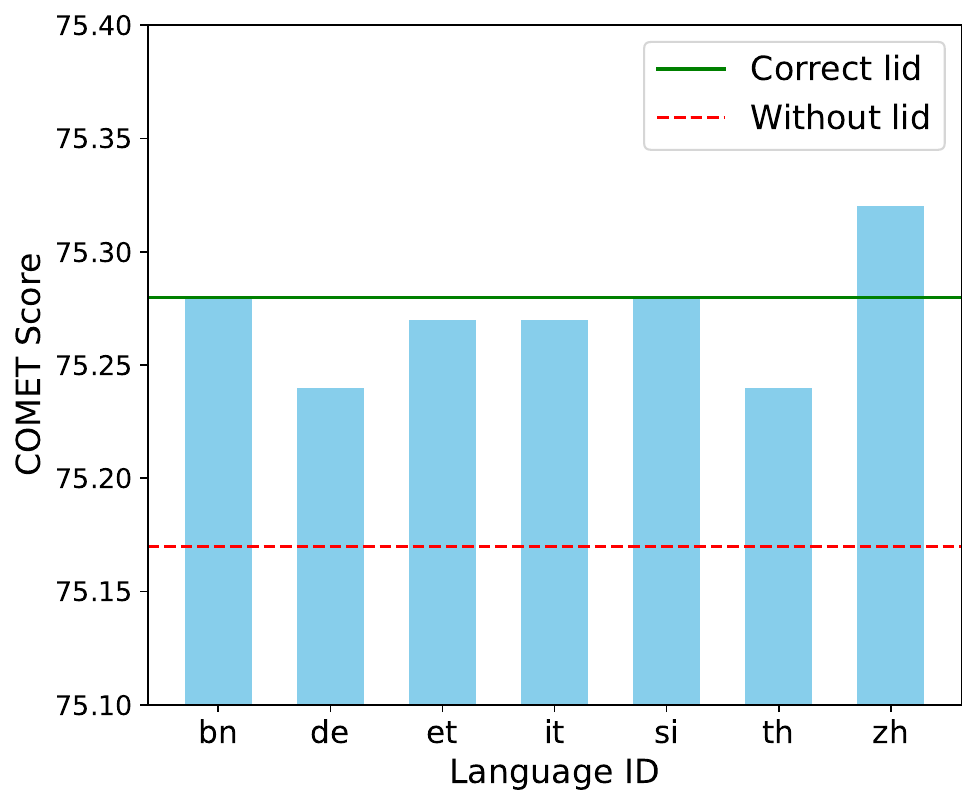}
    \caption{Experiment on "ro$\rightarrow$en" translation. Wrong $lid$ affects little, while lack of $lid$ hurting a lot.}
    \label{fig:wrong-lid}
\end{figure}

Providing Language ID ($lid$) to router seems beneficial to translation performance, according to Table \ref{table:main-ted59} and Table \ref{table:main-opus100}. However, replacing the correct $lid$ with a random one does not degenerate model performance much. Here, we take the translation direction "ro$\rightarrow$en" as an example. The experiment is conducted on Ted-59 dataset, and we replace the $lid$ for the router's input. The results are depicted in Figure \ref{fig:wrong-lid}. The green line and red dashed line denote model with correct $lid$ and without $lid$, while the blue bars denote the wrong $lid$ provided. It proves that having a $lid$ is more critical than having the $lid$ correct. We conjecture this is because the language difference is limited, and the distribution of $\delta$ is limited. However, removing the $lid$ may cause a systematic shift of $\delta$, similar to that discussed in Section \ref{section:systematic-shift-delta}.

In this experiment, the performance differences between different settings are nuanced, and BLEU scores are the same for several. Therefore, we evaluate with the COMET score, which is more accurate and better at showing subtle differences.

\end{document}